\documentclass{article}
\usepackage{spconf,amsmath,graphicx}
\usepackage{CJKutf8} 
\usepackage{comment} 
\usepackage{textcomp}
\usepackage{tikz}
\usepackage{bbding}
\usepackage{pifont}
\usepackage{wasysym}
\usepackage{amssymb}
\usepackage{multirow}

\title{Mitigating the Impact of Speech Recognition Errors on \\Spoken Question Answering by Adversarial Domain Adaptation}
\name{Chia-Hsuan Lee, Yun-Nung Chen, Hung-Yi Lee}
\address{College of Electrical Engineering and Computer Science\\National Taiwan University, Taiwan\\
\small{\texttt{chiahsuan.li@gmail.com} , \texttt{y.v.chen@ieee.org} , \texttt{tlkagkb93901106@gmail.com}}}
\begin{document}
%
\maketitle
\begin{abstract}
Spoken question answering (SQA) is challenging due to complex reasoning on top of the spoken documents.
The recent studies have also shown the catastrophic impact of automatic speech recognition (ASR) errors on SQA. 
Therefore, this work proposes to mitigate the ASR errors by aligning the mismatch between ASR hypotheses and their corresponding reference transcriptions. 
An adversarial model is applied to this domain adaptation task, which forces the model to learn domain-invariant features the QA model can effectively utilize in order to improve the SQA results.
The experiments successfully demonstrate the effectiveness of our proposed model, and the results are better than the previous best model by 2\% EM score.
\end{abstract}
\section{Introduction}
\label{sec:intro}

Question answering (QA) has drawn a lot of attention in the past few years. 
QA tasks on images \cite{zitnick2016adopting} have been widely studied, but most focused on understanding text documents~\cite{rajpurkar2016squad}. 
A representative dataset in text QA is SQuAD~\cite{rajpurkar2016squad}, in which several end-to-end neural models have accomplished promising performance~\cite{yu2018qanet}. 
Although there is a significant progress in machine comprehension (MC) on text documents, MC on spoken content is a much less investigated field. 
In spoken question answering (SQA), after transcribing spoken content into text by automatic speech recognition (ASR), typical approaches use information retrieval (IR) techniques~\cite{shiang2014spoken} to find the proper answer from the ASR hypotheses. 
One attempt towards QA of spoken content is TOEFL listening comprehension by machine~\cite{tseng2016towards}. 
TOEFL is an English examination that tests the knowledge and skills of academic English for English learners whose native languages are not English. 
Another SQA corpus is Spoken-SQuAD\cite{li2018spoken}, which is automatically generated from SQuAD dataset through Google Text-to-Speech (TTS) system. 
Recently ODSQA, a SQA corpus recorded by real speakers, is released~\cite{lee2018odsqa}. 

To mitigate the impact of speech recognition errors, using sub-word units is a popular approach for speech-related downstream tasks.
It has been applied to spoken document retrieval~\cite{ng1997subword} and spoken term detection~\cite{van2017constructing}
The prior work showed that, using phonectic sub-word units brought improvements for both Spoken-SQuAD and ODSQA~\cite{li2018spoken}.

Instead of considering sub-word features, this paper proposes a novel approach to mitigate the impact of ASR errors.
We consider reference transcriptions  and ASR hypotheses as two domains, and
adapt the source domain data (reference transcriptions) to the target domain data (ASR hypotheses) by projecting these two domains in the shared common space.
Therefore, it can effectively benefit the SQA model by improving the robustness to ASR errors in the SQA model.

Domain adaptation has been successfully applied on computer vision~\cite{ganin2016domain} and speech recognition~\cite{shinohara2016adversarial}. 
It is also widely studied on NLP tasks such as sequence tagging and parsing~\cite{yang2017transfer,mcclosky2010automatic,chiticariu2010domain}. 
Recently, adversarial domain adaptation has already been explored on spoken language understanding (SLU).
Liu and Lane learned domain-general features to benefit from multiple dialogue datasets~\cite{liu2017multi}; Zhu et al. learned to transfer the model from the transcripts side to the ASR hypotheses side~\cite{zhu2018robust}; Lan et al. constructed a shared space for slot tagging and language model~\cite{lan2018semi}.
This paper extends the capability of adversarial domain adaptation for SQA, which has not been explored yet. 

\begin{figure*}[t]
  \vspace{-3mm}
  \centering
  \includegraphics[width=.96\linewidth]{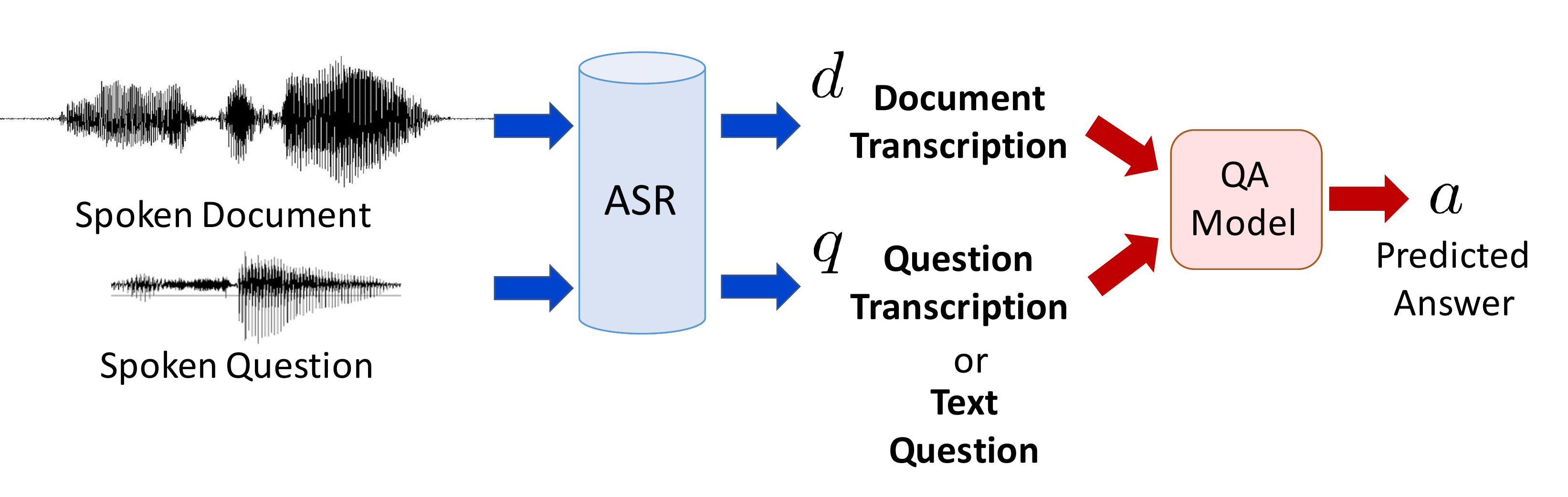}
  \vspace{-5mm}
  \caption{Flow diagram of the SQA system. }
  \label{fig:overview}

\end{figure*}

\section{Spoken Question Answering}

In SQA, each sample is a triple, $(q, d, a)$, where $q$ is a \emph{question} in either spoken or text form, $d$ is a \emph{multi-sentence spoken-form document}, and $a$ is the \emph{answer} in text from.
The task of this work is extractive SQA; that means $a$ is a word span from the reference transcription of $d$.
An overview framework of SQA is shown in Figure~\ref{fig:overview}.
In this paper, we frame the source domain as reference transcriptions and the target domain as ASR hypotheses.
Hence, we can collect source domain data more easily, and adapt the model to the target domain.

In this task, when the machine is given a spoken document, it needs to find the answer of a question from the spoken document. 
SQA can be solved by the concatenation of an ASR module and a question answering module.
Given the ASR hypotheses of a spoken document and a question, the question answering module can output a text answer. 

The most intuitive way to evaluate the text answer is to directly compute the \textbf{Exact Match (EM)} and \textbf{Macro-averaged F1 scores (F1)} between the predicted text answer and the ground-truth text answer.
We used the standard evaluation script from SQuAD~\cite{rajpurkar2016squad} to evaluate the performance.

\section{Question Answering Model}
\label{sec:QA model}

The used architecture of the QA model is briefly summarized below.
Here we choose QANet~\cite{yu2018qanet} as the base model due to the following reasons:
1) it achieves the second best performance on SQuAD, and
2) since there are completely no recurrent networks in QANet, its training speed is 5x faster than BiDAF~\cite{seo2016bidirectional} when reaching the same performance on SQuAD.

\begin{figure}[t]
  \centering
  \includegraphics[width=\linewidth]{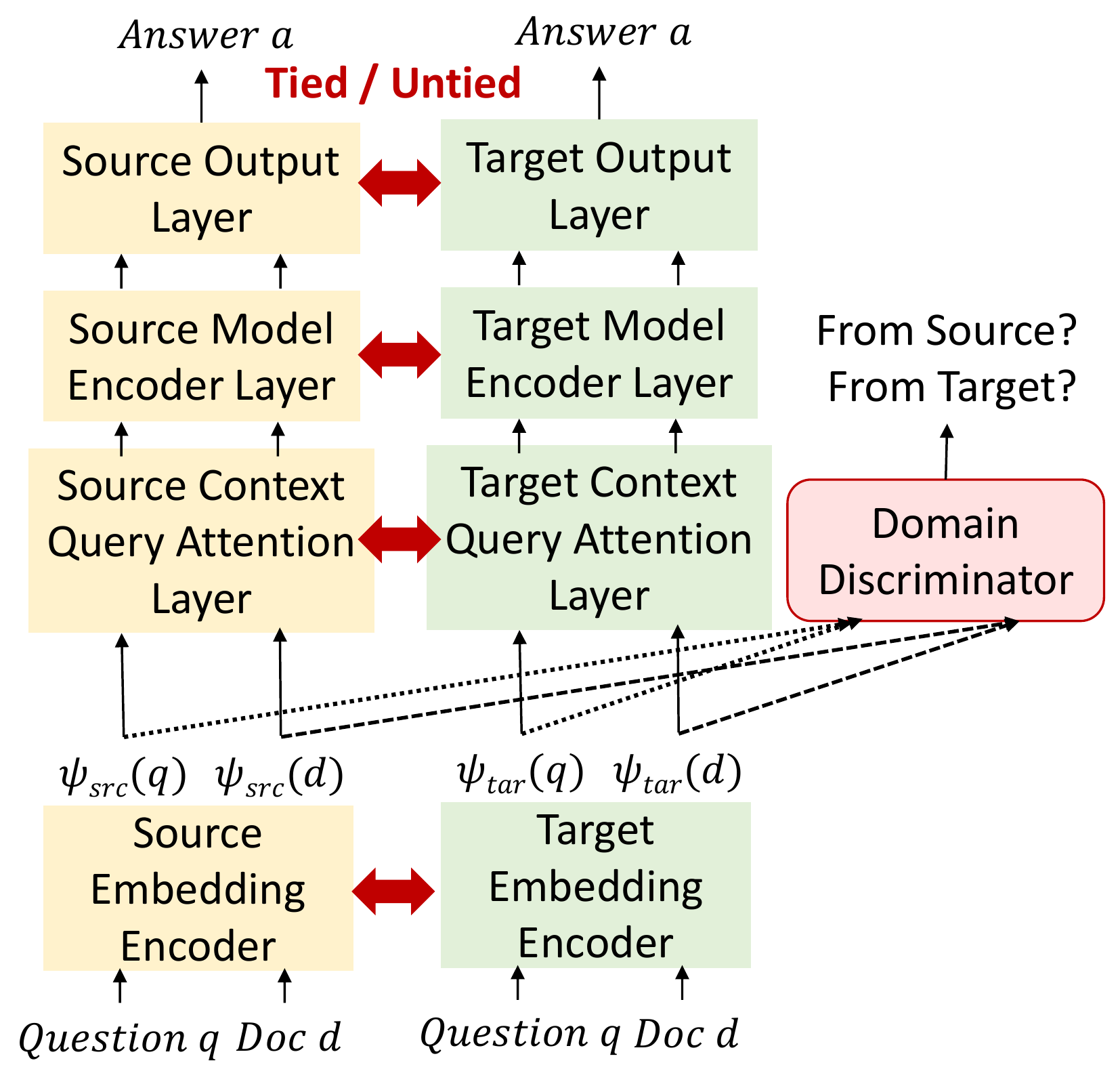}
  \caption{The overall architecture of the proposed QA model with a domain discriminator. 
  Each layer can be tied or untied between the source and target models.}
  \label{fig:GAN}
\end{figure}

The network architecture is illustrated in Figure~\ref{fig:GAN}.
The left blocks and the right blocks form two QANets, each of which takes a document and a question as the input and outputs an answer.
In QANet, firstly, an \emph{embedding encoder} obtains word and character embeddings for each word in $q$ or $d$ and then models the temporal interactions between words and refines word vectors to contextualized word representations. 
All encoder blocks used in QANet are composed exclusively of depth-wise separable convolutions and self-attention.
The intuition here is that convolution components can model local interactions and self-attention components focus on modeling global interactions. 
The \emph{context-query attention layer} generates the question-document similarity matrix and computes the question-aware vector representations of the context words. 
After that, a \emph{model encoder layer} containing seven encoder blocks captures the interactions among the context words conditioned on the question. Finally, the \emph{output layer} predicts a start position and an end position in the document to extract the answer span from the document. 

\section{Domain Adaptation Approach}
\label{sec:pagestyle}
The main focus of this paper is to apply domain adaptation for SQA.
In this approach, we have two SQA models (QANets), one trained from  target domain data (ASR hypotheses) and another trained from source domain data (reference transcriptions).
Because the two domains share common information, some layers in these two models can be tied in order to model the shared features. 
Hence, we can choose whether each layer in the QA model should be shared. 
Tying the weights between the source layer and the target layer in order to learn a symmetric mapping is to project both source and target domain data to a shared common space. 
Different combinations will be investigated in our experiments.

More specifically, we incorporate a domain discriminator into the SQA model shown in Figure~\ref{fig:GAN}, which can enforce the embedding encoder to project the sentences from both source and target domains into a shared common space and consequentially to be ASR-error robust.
Although the embedding encoder for both domains may \emph{implicitly} learn some common latent representations,  adversarial learning can provide a more \emph{direct} training signal for aligning the output distribution of the embedding encoder  from both domains.
The embedding encoder takes in a sequence of word vectors and generates a sequence of hidden vectors with the same length.
We use $\Psi_\text{tar}(q)$ and $\Psi_\text{tar}(d)$ ($\Psi_\text{src}(q)$ and $\Psi_\text{src}(d)$) to represent the hidden vector sequence given the question $q$ and the document $d$ in the target (source) domain respectively.

The domain discriminator $D$ focuses on identifying the domain of the vector sequence is from given $\Psi_\text{tar}$ or $\Psi_\text{src}$, where the objective is to minimize $L_\text{dis}$.
\begin{eqnarray}
\label{eq2}
L_\text{dis}  = E_{(q,d,a) \sim \text{tar}}  \,[\log D(\Psi_\text{tar}(q))+\log D(\Psi_\text{tar}(d))] \\
+ E_{(q,d,a) \sim \text{src}} \, [\log (1 -  D(\Psi_\text{src}(q))+\log (1 -  D(\Psi_\text{src}(d))]. \nonumber
\end{eqnarray}
Given a training example from the target domain ($(q,d,a) \sim \text{tar}$), $D$ learns to assign a lower score to $q$ and $d$ in that example, that is, to minimize $D(\Psi_\text{tar}(q))$ and $D(\Psi_\text{tar}(d))$.  
On the other hand, given a training example from the source domain ($(q,d,a) \sim \text{src}$), $D$ learns to assign a larger value to $q$ and $d$.

Furthermore, we update the parameters of the embedding encoders to maximize the domain classification loss $L_\text{dis}$, which works adversarially towards the domain discriminator. 
We thus expect the model to learn features and structures that can generalize across domains when the outputs of $\Psi_\text{src}$ are indistinguishable from the outputs of $\Psi_\text{tar}$. 
The loss function for embedding encoder, $L_\text{enc}$, is formulated as
\begin{equation}
    L_\text{enc} =  L_\text{qa} - \lambda_G L_\text{dis}, \label{eq:L_pri}
\end{equation}
where $\lambda_G$ is a hyperparameter.
The two embedding encoders in the QA model are learned to maximize $L_\text{dis}$ while minimizing the loss for QA, $L_\text{qa}$.
Because the parameters of other layers in QA model are independent to the loss of the domain discriminator, the loss function of other layers, $L_\text{other}$, is equivalent to  $L_\text{qa}$, that is, $L_\text{other} =  L_\text{qa}$.

Although the discriminator is applied to the output of embedding encoder in Figure~\ref{fig:GAN}, it can be also applied to other layers.\footnote{In the  experiments, we found that applying the domain discriminator to embedding encoders yielded the best performance.}
Considering that almost all QA model contains such embedding encoders, 
the proposed approach is expected to generalize to other QA models in addition to QANet.

\section{Experiments}

\subsection{Corpus}

Spoken-SQuAD is chosen as the target domain data for training and testing. Spoken-SQuAD~\cite{li2018spoken} is an automatically generated corpus in which the document is in spoken form and the question is in text form.
The reference transcriptions are from SQuAD~\cite{rajpurkar2016squad}.
There are 37,111 and 5,351 question answer pairs in the training and testing sets respectively, and the word error rate (WER) of both sets is around 22.7\%.

The original SQuAD, Text-SQuAD, is chosen as the source domain data, where only question answering pairs appearing in Spoken-SQuAD are utilized. 
In our task setting, during training we train the proposed QA model on both Text-SQuAD and Spoken-SQuAD training sets.
While in the testing stage, we evaluate the performance on Spoken-SQuAD testing set.

\subsection{Experiment Setup} 

We utilize \texttt{fasttext}~\cite{bojanowski2016enriching} to generate the embeddings of all words from both Text-SQuAD and Spoken-SQuAD. 
We adopt the phoneme sequence embeddings to replace the original character sequence embeddings using the method proposed by Li et al. \cite{li2018spoken}.
The source domain model and the target domain model share the same set of word embedding matrix to improve the alignment between these two domains.  

W-GAN is adopted for our domain discriminator~\cite{gulrajani2017improved}, which stacks 5 residual blocks of 1D convolutional layers with 96 filters and filter size 5 followed by one linear layer to convert each input vector sequence into one scalar value. 

All models used in the experiments are trained with batch size 20, using \texttt{adam} with learning rate $1e-3$ and the early stop strategy.
The dimension of the hidden state is set to 96 for all layers, and the number of self-attention heads is set to 2.
The setup is slightly different but better than the setting suggested by the original QAnet.

\subsection{Results}

\begin{table}[t!]
\centering
\caption{Illustration of domain mismatch, where the models are trained on the source domain (Text-SQuAD; T-SQuAD) or the target domain (Spoken-SQuAD; S-SQuAD) and then evaluated on both source and target domains.}
\label{tab:domain}
\vspace{2mm}
\begin{tabular}{|c|c|c|c|c|c|}
\hline
\multicolumn{2}{|c|}{\bf Model}  &
\multicolumn{2}{|c|}{\bf T-SQuAD} &
\multicolumn{2}{|c|}{\bf S-SQuAD} \\
\cline{3-6}
\multicolumn{2}{|c|}{\bf Training} &\textbf{EM} & \textbf{F1} &\textbf{EM} & \textbf{F1} \\
\hline
\hline
T-SQuAD & (a) & 61.31 & 72.66 & 42.27 & 55.61 \\
S-SQuAD  & (b) & 45.52 & 57.39 & 48.93 & 61.20\\
\hline
Finetune & (c) & 54.83 & 66.45 & 49.60 & 61.85 \\
\hline
\end{tabular}
\end{table}

\subsubsection{Domain Mismatch}
First, we highlight the domain mismatch phenomenon in our experiments shown in Table~\ref{tab:domain}.
Row (a) is when QANet is trained on Text-SQuAD, row (b) is when QANet is trained on Spoken-SQuAD, and row (c) is when QANet is trained on Text-SQuAD and then finetuned on Spoken-SQuAD. 
The columns show the evaluation on the testing sets of Text-SQuAD and Spoken-SQuAD. 

It is clear that the performance drops a lot when the training and testing data mismatch, indicating that model training on ASR hypotheses can not generalize well on reference transcriptions.
The performance gap is nearly 20\% F1 score (72\% to 55\%).
The row (c) shows the improved performance when testing on S-SQuAD due to the transfer learning via fine-tuning.

\subsubsection{Effectiveness of Adversarial Domain Adaptation}

\begin{table}[t!]
\centering
\caption{The EM/F1 scores of proposed adversarial domain adaptation approaches over Spoken-SQuAD.}
\label{tab:GAN}
\vspace{2mm}
\begin{tabular}{|lccc|}
\hline
\multicolumn{2}{|c|}{{\textbf{Model}}}  & \textbf{EM} & \textbf{F1}  \\
\hline
\hline
\emph{Baseline} & & &\\
~~~~~S-SQuAD & (a) & 48.93 & 61.20 \\
~~~~~Finetune & (b) & 49.60 & 61.85 \\
~~~~~Li et al.~\cite{li2018spoken} & (c)  & 49.07 & 61.16 \\
\hline
\emph{Adverarial} & & &\\
~~~~~Lan et al.~\cite{lan2018semi} & (d) & 49.13 & 61.80 \\
~~~~~Completely Shared & (e) & 49.57 & 61.48 \\
~~~~~(e) + GAN on Embedding & (f) & \textbf{51.10} & \textbf{63.11} \\
~~~~~(e) + GAN on Attention & (g) & 48.30 & 61.11 \\
\hline
\end{tabular}
\end{table}

To better demonstrate the effectiveness of the proposed model, we compare with baselines and show the results in Table~\ref{tab:GAN}. 
The baselines are: (a) trained on S-SQuAD, (b) trained on T-SQuAD and then fine-tuned on S-SQuAD, and (c) previous best model trained on S-SQuAD~\cite{li2018spoken} by using Dr.QA~\cite{chen2017reading}.
We also compare to the approach proposed by Lan et al.~\cite{lan2018semi} in the row (d).
This approach is originally proposed for spoken language understanding, and we adopt the same approach on the setting here.
The approach models domain-specific features from the source and target domains separately by two different embedding encoders with a shared embedding encoder for modeling domain-general features.
The domain-general parameters are adversarially trained by domain discriminator. 

Row (e) is the model that the weights of all layers are tied between the source domain and the target domain. 
Row (f) uses the same architecture as row (e) with an additional domain discriminator applied to the embedding encoder.
It can be found that row (f) outperforms row (e), indicating that the proposed domain adversarial learning is helpful. Therefore, our following experiments contain domain adversarial learning. 
The proposed approach (row (f)) outperforms previous best model (row (c)) by 2\% EM score and over 1.5\% F1 score.
We also show the results of applying the domain discriminator to the top of context query attention layer in row (g), which obtains poor performance.
To sum it up, incorporating adversarial learning by applying the domain discriminator on top of the embedding encoder layer is effective. 


\subsubsection{Which Layer to Share?}
Layer weight tying or untying within the model indicates different levels of symmetric mapping between the source and target domains.
Different combinations are investigated and shown in Table~\ref{tab:share_comparison}. 
The row (a)  in which all layers are tied is the row (e) of Table~\ref{tab:GAN}. 
The results show that untying context-query attention layer L2 (rows (c, f, g)) or model encoder layer L3 (rows (d, f, h)) lead to degenerated solutions in comparison to row (a) where all layers are tied. 
Untying both of them simultaneously leads to the worst performance which is even worse than the finetuning (row (g) v.s. (c) from Table~\ref{tab:GAN}). 
These results imply that sharing the \emph{context-query attention layer} and the \emph{model encoder layer} are important for domain adaptation on SQA. 
We conjecture that these two layers benefit from training on source domain data where there are no ASR errors, so the QA model learns to conduct attention or further reason well on target domain data with ASR errors.

Overall, it is not beneficial to untie any layer, because no information can be shared across different domains.
Untying the embedding encoder L1 and the output layer L4 leads to the least degradation in comparison to row (a). 

\begin{table}[t!]
\centering
\caption{Investigation of different layer tying mechanisms, where \checkmark means that weights of the layer are tied between the source model and the target model. (L1: embedding encoder, L2: context query attention layer, L3: model encoder layer, L4: output layer.)}
\vspace{2mm}
\label{tab:share_comparison}
\begin{tabular}{|c |c c c c | c c|}
\hline
\multicolumn{1}{|c|}{{\textbf{Combination}}} & \bf L1 & \bf L2 & \bf L3 & \bf L4 &\textbf{EM} & \textbf{F1}  \\
\hline
\hline
 (a) & \checkmark &\checkmark & \checkmark  & \checkmark  & \textbf{51.10} & \textbf{63.11} \\
 (b) & - & \checkmark & \checkmark & \checkmark & 50.25 & 62.41 \\
 (c) & - & - & \checkmark & \checkmark  & 49.72 & 61.97 \\
 (d) & - & \checkmark & - & \checkmark & 48.83 & 61.80 \\
 (e) & - & \checkmark & \checkmark & - & 51.09 & 62.97 \\
 (f) & \checkmark & - & - & \checkmark & 49.01 & 61.40 \\
 (g)  & \checkmark & - & \checkmark &  - & 49.28 & 61.71 \\
 (h)  & \checkmark & \checkmark & - & -  & 49.61 & 61.72 \\
\hline
\end{tabular}
\end{table}

\section{Conclusion}
In this work, we incorporate a domain discriminator to align the mismatched domains between ASR hypotheses and reference transcriptions. The adversarial learning allows the end-to-end QA model to learn domain-invariant features and improve the robustness to ASR errors.
The experiments demonstrate that the proposed model successfully achieves superior performance and outperforms the previous best model by 2\% EM score and over 1.5\% F1 score.
\bibliographystyle{IEEEbib}
\bibliography{strings,refs}

\end{document}